\documentclass[11pt]{article}

\usepackage[preprint]{acl}

\usepackage{times}
\usepackage{latexsym}

\usepackage[T1]{fontenc}

\usepackage[utf8]{inputenc}

\usepackage{microtype}

\usepackage{inconsolata}

\usepackage{graphicx}
\usepackage[table]{xcolor}
\usepackage{multirow}
\usepackage{booktabs}
\usepackage{array}
\usepackage{pifont}
\usepackage{subcaption}
\usepackage{makecell}
\usepackage{diagbox}

\usepackage{enumitem}

\title{EgoEsportsQA: An Egocentric Video Benchmark for Perception and Reasoning in Esports}

\author{Jianzhe Ma$^1$\thanks{Equal contribution} ~~~ Zhonghao Cao$^{2}$\footnotemark[1] ~~~ Shangkui Chen$^1$
\\\textbf{Yichen Xu}$^1$ ~~~ \textbf{Wenxuan Wang}$^{1}$\footnotemark[2] ~~~ \textbf{Qin Jin}$^1$\thanks{Corresponding authors} \\
$^1$Renmin University of China\\
$^2$Beijing University of Posts and Telecommunications\\
\texttt{\{majianzhe, wangwenxuan, qjin\}@ruc.edu.cn}
}

\begin{document}
\maketitle
\begin{abstract}
While video large language models (Video-LLMs) excel in understanding slow-paced, real-world egocentric videos, their capabilities in high-velocity, information-dense virtual environments remain under-explored. Existing benchmarks focus on daily activities, yet lack a rigorous testbed for evaluating fast, rule-bound reasoning in virtual scenarios. To fill this gap, we introduce EgoEsportsQA, a pioneering video question-answering (QA) benchmark for grounding perception and reasoning in expert esports knowledge. We curate 1,745 high-quality QA pairs from professional matches across 3 first-person shooter games via a scalable six-stage pipeline. These questions are structured into a two-dimensional decoupled taxonomy: 11 sub-tasks in the cognitive capability dimension (covering perception and reasoning levels) and 6 sub-tasks in the esports knowledge dimension. Comprehensive evaluations of state-of-the-art Video-LLMs reveal that current models still fail to achieve satisfactory performance, with the best model only 71.58\%. The results expose notable gaps across both axes: models exhibit stronger capabilities in basic visual perception than in deep tactical reasoning, and they grasp overall macro-progression better than fine-grained micro-operations. Extensive ablation experiments demonstrate the intrinsic weaknesses of current Video-LLM architectures. Further analysis suggests that our dataset not only reveals the connections between real-world and virtual egocentric domains, but also offers guidance for optimizing downstream esports applications, thereby fostering the future advancement of Video-LLMs in various egocentric environments.
\end{abstract}

\section{Introduction}
\label{sec:intro}

\begin{figure*}[ht]
  \centering
  \includegraphics[width=0.95\linewidth]{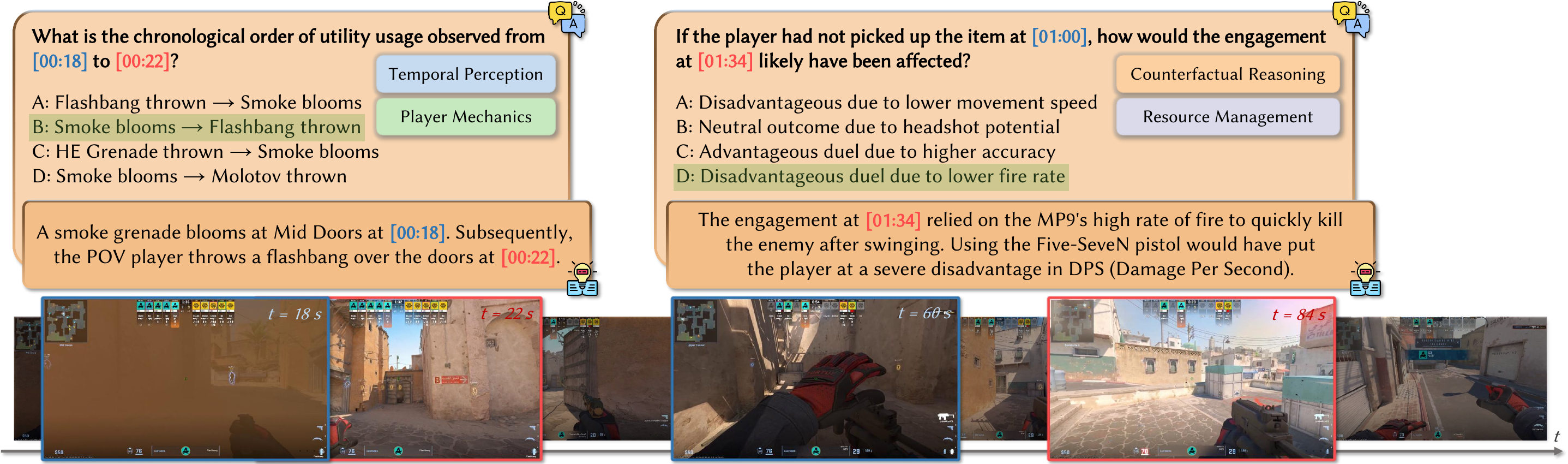}
  \caption{Examples from EgoEsportsQA. The benchmark requires high-frequency visual perception (left) and expert tactical reasoning (right). Each question has a time anchor for target events, organized by our two-dimensional taxonomy.}
  \vspace{-0.5cm}
  \label{fig:figure1_example}
\end{figure*}

The pursuit of Artificial General Intelligence (AGI) necessitates agents capable of perceiving, reasoning, and acting in complex and dynamic environments. Recent advances in Multimodal Large Language Models (MLLMs)~\cite{achiam2023gpt4, liu2023llava} and their video-centric counterparts, Video Large Language Models (Video-LLMs)~\cite{zhang2023videollama, li2024llavanextvideo}, have demonstrated strong perception and reasoning abilities across diverse visual benchmarks, laying a solid foundation for multimodal world understanding. However, while these models excel in real-world, relatively slow-paced video understanding tasks~\cite{fan2019egovqa, wong2022assistq, grauman2022ego4d}, their performance remains largely under-explored in \textbf{virtual, high-velocity, and adversarial environments}~\cite{xu2025lmsportssurvey}. The current lack of specialized benchmarks for such settings limits a thorough understanding of MLLMs’ true potential in rapid information processing and complex strategic decision-making.

Esports, as a rapidly growing global industry, presents an ideal testbed for this challenge~\cite{hamari2017esports, reitman2020esports}. Within this domain, \textbf{First-Person Shooter (FPS)} esports stand out due to their high-speed dynamics, intricate strategies, and intense adversarial competition~\cite{banyai2019psychology, jang2020antecedents}. Understanding competitive FPS requires models to master a hierarchical set of abilities: 1) fine-grained visual \textbf{perception} from a constrained first-person field-of-view; 2) deep \textbf{reasoning} that incorporates expert-level tactical knowledge; 3) rapid \textbf{decision-making} based on the above. The integrated capability of perception, reasoning and decision offers a critical perspective for assessing the core competencies of Video-LLMs~\cite{zhao2025dlsportssurvey}. Notably, while real-world egocentric videos are costly to collect~\cite{damen2018scaling, grauman2022ego4d}, high-quality first-person footage can be efficiently obtained from professional esports matches, providing a valuable resource for constructing visual intelligence benchmarks.

However, existing research exhibits a gap in evaluating such capabilities. In \textbf{esports}, most efforts focus either on mining structured data~\cite{xenopoulos2022analyzing, hirota2024predicting} or on downstream applications such as commentary generation~\cite{tanaka2021lol, zhang2022moba} and game-playing agents~\cite{ma2024large, ma2025ava}, yet fail to systematically quantify models’ core perception and reasoning abilities. Meanwhile, research in \textbf{egocentric video understanding} mainly targets daily activities in cooperative environments~\cite{mangalam2023egoschema, ye2025mmego}, lacking the adversarial pressure, information-dense User Interface (UI) elements, and rule-bound tactical reasoning inherent to FPS.

To address this gap, we introduce \textbf{EgoEsportsQA}---the first video question-answering benchmark for FPS esports from an egocentric perspective, as shown in Figure~\ref{fig:figure1_example}. 
Built upon high-quality recordings from professional tournaments, EgoEsportsQA is constructed via a scalable \textbf{six-stage pipeline} and annotated by experienced annotators with strong domain knowledge. It comprises \textbf{1,745} carefully curated QA pairs derived from \textbf{364} video clips across \textbf{3} popular FPS titles. Each QA pair in the benchmark is categorized along two orthogonal dimensions: the \textbf{cognitive capability} dimension with 11 sub-tasks spanning perception and reasoning levels, and the \textbf{esports knowledge} dimension with 6 sub-tasks covering macro-progression and micro-operation categories.
Additionally, EgoEsportsQA employs time anchor alignment to eliminate ambiguous reasoning over long videos, enabling fine-grained temporal grounding and precise answer inference. Meanwhile, we neutralize textual shortcuts to enforce strict visual dependency, yielding text-only performance near random guessing and ensuring evaluation relies on genuine visual understanding.

We systematically evaluate \textbf{9} representative Video-LLMs on EgoEsportsQA, including both proprietary~\cite{pichai2025gemini3, openai2025gpt5, anthropic2025sonnet45, bytedance2025seed18} and open-source~\cite{li2024llavanextvideo, li2025llavaonevision, bai2025qwen3, wang2025internvl35, yang2025egolife} models. 
Experimental results demonstrate that even state-of-the-art models achieve only \textbf{71.58\%} accuracy.
Notably, models perform worse on tactical reasoning tasks than on basic perception tasks, and struggle more with fine-grained micro-operation than macro-progression, exposing the \textbf{gap between ``seeing'' and ``understanding''} as well as their \textbf{deficiency in modeling high-speed dynamic visual content}.
Ablation studies validate the necessity of visual and audio input, as well as suitable frame sampling rates and resolutions, for effective egocentric video understanding and reveal the challenges of long-video modeling in our benchmark.
Further exploratory experiments illustrate the strong \textbf{virtual-to-real transferability}, demonstrating our benchmark’s potential as a valuable evaluation \textbf{platform for downstream applications} and highlighting its significance in developing Video-LLMs with more robust egocentric understanding capabilities.

In summary, our contributions are as follows:
\begin{itemize}[leftmargin=*, topsep=2pt]
\item We initiate the study of expert-level egocentric understanding in high-velocity, competitive FPS esports scenarios. To this end, we construct \textbf{EgoEsportsQA}, a large-scale benchmark with 1,745 carefully curated QA pairs built via a scalable six-stage pipeline.
\item Through systematic evaluation, we identify bottlenecks in current Video-LLMs, particularly their difficulty in transitioning from visual perception to tactical reasoning, as well as their deficiency in fine-grained micro-operation understanding.
\item We further validate strong virtual-to-real transferability, demonstrating the benchmark's value as a diagnostic evaluation platform for downstream tasks and its potential to facilitate the development of more capable egocentric Video-LLMs.
\end{itemize}

\section{Related Work}
\label{sec:related_work}

\paragraph{\textbf{Esports-Related Research.}}
As a domain characterized by high-speed dynamics and complex strategies, esports has attracted increasing research interest. Existing works explore both data-driven analytics and multimodal task execution. Data-driven tasks such as win probability prediction~\cite{xenopoulos2022analyzing, hirota2024predicting, hayakawa2025round}, player behavior analysis~\cite{durst2024learning, wang2025x}, event detection~\cite{ringer2019multimodal, ng20253m}, and data classification~\cite{jubaer2024analyzing, lu2023mug} mainly leverage structured game logs, images, or videos~\cite{xenopoulos2022esta, bialecki2023sc2egset, xu2023cs, duffy2025cdops}. Multimodal tasks such as commentary generation~\cite{tanaka2021lol, zhang2022moba, xu2022toward, mamoru2022conceptual, wang2024commentary, yu2025single} and esports-playing agents~\cite{ma2024large, ma2024adaptive, ma2025ava, wang2025data} utilize broadcast video, audio, or game state to produce human-like outputs. Though progress has been made, these works either bypass visual perception entirely or evaluate vision models on end-to-end tasks without isolating core perception and reasoning abilities. 
In contrast, \textbf{EgoEsportsQA} fills this gap with a video QA benchmark that assesses these core capabilities using egocentric video from professional FPS matches, bridging basic multimodal understanding and complex downstream applications.

\paragraph{\textbf{Egocentric Video Understanding.}}
Egocentric video understanding has evolved from short-term action recognition~\cite{damen2018scaling} to long-term activity planning~\cite{mangalam2023egoschema, yang2025egolife}, pushing video length from seconds to hours. Recent benchmarks extend egocentric understanding beyond indoor daily activities~\cite{fan2019egovqa, jia2022egotaskqa} to instructional tasks~\cite{wong2022assistq}, surgery, sports~\cite{li2025egocross}, and even cross-view (first- and third-person) settings~\cite{he2025egoexobench}. These approaches primarily focus on understanding real-world daily activities or procedural skills in predictable, cooperative environments. 
In contrast, FPS esports offers a representative virtual egocentric setting with high-speed dynamics, information-dense UI elements, and adversarial tactical reasoning. It also serves as an ideal testbed for investigating the connection and generalization between real-world and virtual egocentric scenarios. As shown in Table~\ref{tab:ego_centric_comparison}, existing datasets primarily cover daily and real-world domains, leaving the fast-paced virtual esports landscape largely unexplored.

\begin{table}[t]
\centering

\resizebox{\columnwidth}{!}{
\begin{tabular}{lccrrr}
\toprule
\textbf{Benchmarks} & \textbf{Domain} & \textbf{Anno.} & \textbf{\#Clips}& \textbf{\#QAs} & \textbf{Len.} \\ 
\midrule
EgoVQA~\citeyearpar{fan2019egovqa} & Indoor, Daily & M & $\sim$120 & $\sim$120 & $\sim$60s \\
EgoTaskQA~\citeyearpar{jia2022egotaskqa} & Indoor, Daily & A\&M & $\sim$ 400 & $\sim$ 8,000 & 25s \\
AssistQ~\citeyearpar{wong2022assistq} & Instructional & M & 20 & 106 & 115s \\
EgoSchema~\citeyearpar{mangalam2023egoschema} & Human Activity & A\&M & 5,063 & 5,063 & 180s \\
EgoLifeQA~\citeyearpar{yang2025egolife} & Daily Life & A\&M & 644 & 3,000 & $\sim$4.3h \\
EgoMemoria~\citeyearpar{ye2025mmego} & Daily, Instructional & A & 629 & 7,026 & - \\
EgoCross~\citeyearpar{li2025egocross} & Surgery, Industry, Sports & A\&M & 798 & 957 & 22.5s \\
\midrule
\rowcolor{gray!15} \textbf{EgoEsportsQA} & FPS Esports & A\&M & 364 & 1,745 & 73.4s \\
\bottomrule
\end{tabular}%
}
\caption{Comparison with representative egocentric video QA benchmarks, including task domain, annotation method (M: manual, A: automatic), number of clips (\#Clips), number of QAs (\#QAs), and average clip duration (Len.).}
\vspace{-0.5cm}
\label{tab:ego_centric_comparison}
\end{table}

\paragraph{\textbf{Video Large Language Models.}}
Research in Video Large Language Models (Video-LLMs) has achieved significant progress in recent years. Proprietary models such as Gemini 3~\cite{pichai2025gemini3} and GPT-5~\cite{openai2025gpt5} demonstrate strong performance on complex video understanding tasks, while open-source alternatives like LLaVA-One\-Vision~\cite{li2025llavaonevision}, LLaVA-Video~\cite{zhang2025llavavideo}, and Qwen3-VL~\cite{bai2025qwen3} achieve competitive results on general video understanding benchmarks such as MVBench~\cite{li2024mvbench} and VideoMME~\cite{fu2025videomme}. 
However, these models often struggle in specialized domains where visual cues are fine-grained and reasoning requires expert-level knowledge.
\textbf{EgoEsportsQA} fills this gap by evaluating Video-LLMs on timestamp-anchored questions grounded in dynamic FPS videos, assessing their fine-grained perception and strategic reasoning abilities for high-velocity egocentric scenarios, thereby facilitating the development of more capable multimodal agents for both virtual and real-world egocentric applications.


\section{EgoEsportsQA}
\label{sec:dataset_egoesportsqa}

In this section, we present the development of the EgoEsportsQA benchmark, covering its six-stage data construction pipeline and detailed dataset statistics.

\subsection{Dataset Construction}

\begin{figure*}[h]
  \centering
  \includegraphics[width=0.95\linewidth]{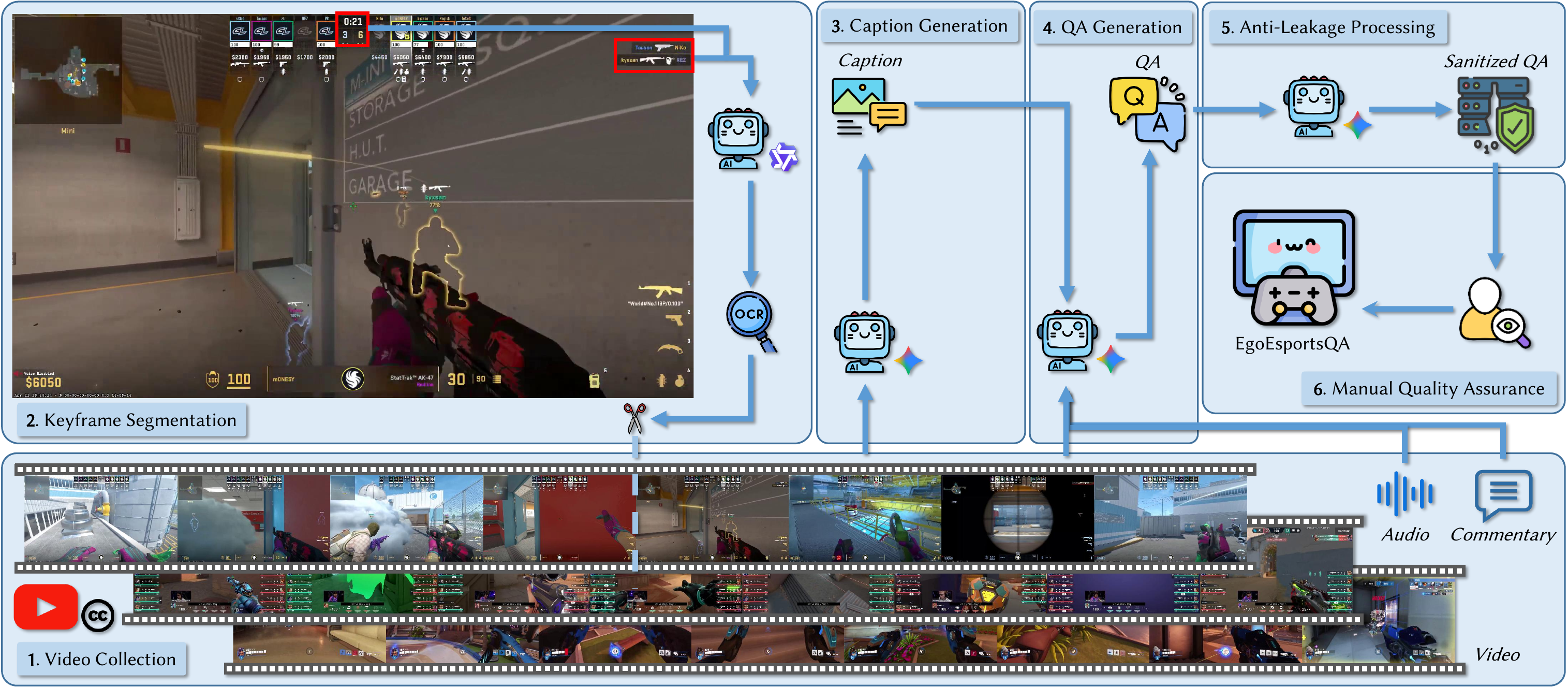}
  \caption{The six-stage data construction pipeline of EgoEsportsQA.}
  \label{fig:figure2_pipeline}
  \vspace{-0.5cm}
\end{figure*}

We construct EgoEsportsQA through a \textbf{scalable six-stage pipeline} designed to ensure diversity, quality, and unbiased evaluation, as shown in Figure~\ref{fig:figure2_pipeline}. The construction details are as follows.

\paragraph{\textbf{Stage 1: Video Collection.}}
To ensure diversity, we select 3 representative FPS titles: \textit{Counter-Strike 2} (CS2)~\cite{valve2023cs2}, \textit{Valorant}~\cite{riot2020valorant}, and \textit{Overwatch 2} (OW2)~\cite{blizzard2022ow2}. We collect first-person perspective recordings from top-tier professional tournaments held between 2023 and 2025, covering 28 distinct maps (totaling 12.3 hours). Formally, each raw esports broadcast is represented as ${V}_{raw} = ({F}_{raw}, {A}_{raw}, {T}_{raw})$, where ${F}_{raw}$ denotes the frame sequence, ${A}_{raw}$ the audio track, and ${T}_{raw}$ the aligned commentary text when available. All videos are sourced from YouTube under \textit{Creative Commons} licenses and are used solely for non-commercial scientific research.

\paragraph{\textbf{Stage 2: Keyframe Segmentation.}}
Raw videos ${V}_{raw}$ are typically long and may contain multiple team fights or rounds. To produce clips that each encapsulate at least one complete team fight---thereby preserving tactical integrity and question complexity---we perform sparse frame sampling at 1 frame per second (fps) and employ Qwen3-VL-8B~\cite{bai2025qwen3} to detect bounding boxes of key UI elements (e.g., timer, scoreboard, and kill feed). 
We then apply an OCR model~\cite{JaidedAI2020easyocr} to extract textual information from these regions, enabling precise identification of key temporal events such as timer resets, score changes, and player deaths. Based on these signals, we segment the original video into a set of tactically coherent clips $\{V_{clip}^{(i)}\}_{i=1}^{M}$, where $\bigcup_{i=1}^{M} V_{clip}^{(i)} \subseteq V_{raw}$. The resulting clip is defined as ${V}_{clip} = ({F}_{clip}, {A}_{clip}, {T}_{clip})$.

\paragraph{\textbf{Stage 3: Caption Generation.}}
These egocentric clips in FPS esports contain rich, information-dense UI elements that are critical for gameplay understanding. To facilitate accurate and high-quality QA generation, we employ Gemini 3 Pro~\cite{pichai2025gemini3} (denoted as $\Phi$) to generate dense frame-level captions. 
Specifically, for each sampled frame $f_t \in {F}_{clip}=\{f_1, f_2, \dots, f_N\}$ (at 1 fps), the model produces a structured caption $C_t = \Phi(f_t)$ that captures both interface states (e.g., health, ammo, minimap) and event-level descriptions (e.g., ``player throws a flashbang towards B site''). This process yields a temporally aligned caption sequence ${C} = \{C_t\}_{t=1}^{N}$.

\paragraph{\textbf{Stage 4: Question-Answer Generation.}}
Based on the multimodal context, we employ Gemini 3 Pro~\cite{pichai2025gemini3} to generate multiple-choice QA pairs for rigorously evaluating Video-LLMs on egocentric esports understanding. We adopt a multiple-choice format to enable objective and scalable evaluation. 
Importantly, each question explicitly incorporates a precise temporal reference (timestamps or time anchors), as shown in Figure~\ref{fig:figure1_example}. This design is motivated by the fact that players may perform visually similar actions at different moments, while the underlying tactical implications can vary over time. 
Therefore, temporal grounding is required for accurate question answering. Formally, the generation process is defined as:
\begin{equation}
    (Q, O, a^*) = \Phi({F}_{clip}, {A}_{clip}, {T}_{clip}, {C})
\end{equation}
where $Q$ is the generated question, $O = \{o_A, o_B, o_C, o_D\}$ denotes the set of candidate options, and $a^* \in O$ is the ground-truth answer, strictly grounded in the video content. A total of 4,926 QA pairs are generated during this stage.

\paragraph{\textbf{Stage 5: Anti-Leakage Processing.}}
To prevent models from exploiting text-based shortcuts or dataset priors (e.g., map-specific locations, hero abilities, or stereotypical tactics) to answer questions correctly under a text-only setting, we perform an anti-leakage sanitization step. 
Specifically, we transform both questions and candidate options into neutralized forms, abstracting away explicit semantic cues that could lead to language-only shortcuts. To ensure a rigorous evaluation, we perform linguistic style harmonization whereby all distractor options are meticulously rewritten to match the sentence structure, length, and technical granularity of the ground-truth $a^*$.
This ensures that all options are linguistically comparable, preventing models from exploiting superficial textual patterns to identify the correct answer. 
This process is defined as:
\begin{equation}
    (\tilde{Q}, \tilde{O}, a^*) = \Phi(Q, O, a^*)
\end{equation}
where $\tilde{Q}$ and $\tilde{O}$ denote the sanitized question and options, respectively.
As shown in Table~\ref{tab:ablation_modality}, statistical analysis indicates that Gemini 3 Flash~\cite{pichai2025gemini3}, when evaluated under a text-only setting, achieves only 24.47\% accuracy on the sanitized QA pairs, which is close to random guessing (25\%)~\cite{chen2024we, fu2025videomme}. This result confirms that the benchmark effectively suppresses textual shortcuts and enforces strong visual dependency.

\paragraph{\textbf{Stage 6: Manual Quality Assurance.}}
To ensure dataset quality, we conduct rigorous manual verification. We recruit 15 annotators with over five years of FPS gaming and spectating experience, and provide systematic training prior to annotation.
Each QA tuple $(\tilde{Q}, \tilde{O}, a^*, {V}_{clip})$ is reviewed in two stages. In the first stage, an annotator refines the question and options based on the video, ensuring 1) linguistic clarity and accurate terminology, and 2) logical consistency across all options, with distractors structurally comparable to the correct answer. In the second stage, another annotator evaluates the overall quality of the QA pair, focusing on question depth and the effectiveness of distractors, and decides whether to retain or discard the sample.
Through this process, we filter out 3,181 low-quality QA pairs, resulting in a final dataset of 1,745 high-quality samples. The high filtering rate ($\sim$65\%) underscores the rigor of our quality control process, prioritizing question depth and reasoning complexity over raw quantity.

\subsection{Dataset Statistics}

\begin{figure*}[h]
  \centering
  \includegraphics[width=0.95\linewidth]{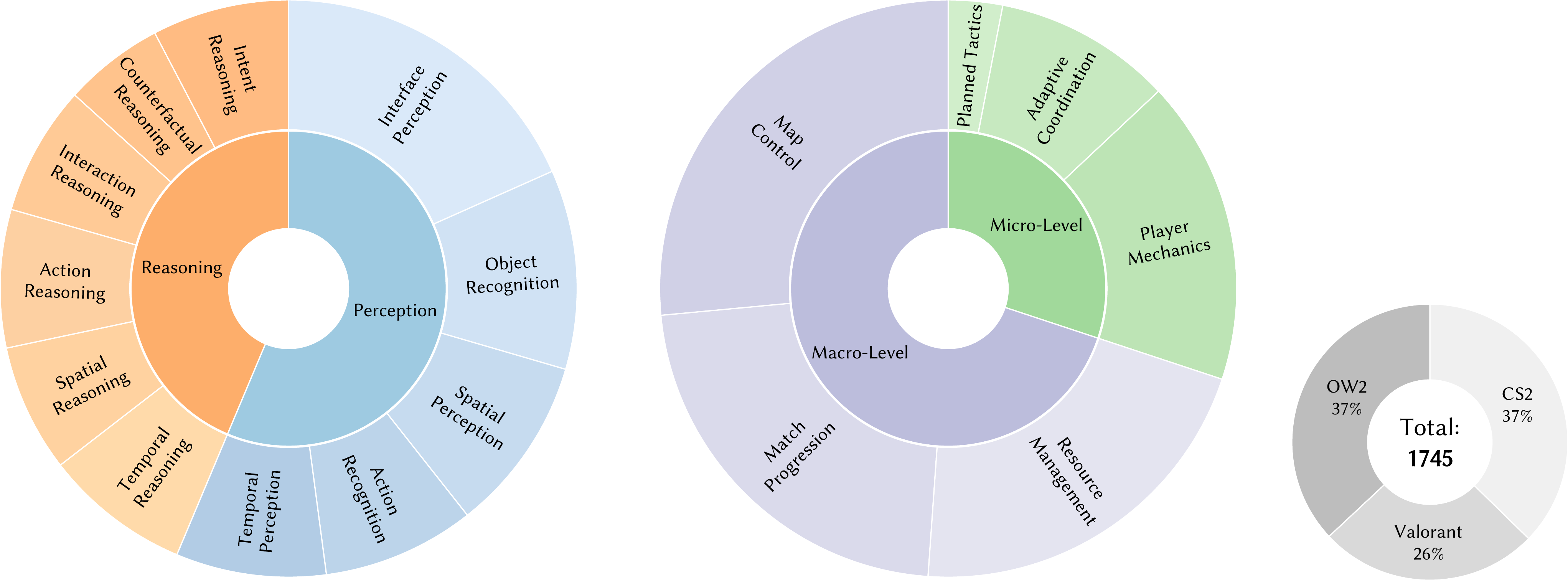}
  \caption{Statistical overview of the EgoEsportsQA benchmark. The dataset is systematically categorized along a two-dimensional decoupled taxonomy: the Cognitive Capability dimension (left) and the Esports Knowledge dimension (middle). The rightmost donut chart illustrates the balanced distribution of the 1,745 QA pairs across 3 representative FPS titles.}
  \vspace{-0.5cm}
  \label{fig:figure3_stat}
\end{figure*}

We provide a detailed statistical overview of EgoEsportsQA to offer a more comprehensive understanding, covering meta information, an orthogonal task taxonomy, and quality control.

\paragraph{\textbf{Meta Information.}}
The benchmark consists of 1,745 high-quality multiple-choice QA pairs derived from 364 unique egocentric video clips from professional esports competition. All clips are provided in 1920$\times$1080 resolution with synchronized audio, and 113 of them include commentary text. The dataset comprises approximately 7.5 hours of footage, with an average clip duration of 73.4 seconds. Most clips fall within the 40 to 100 second range, allowing the dataset to capture complete tactical engagement cycles rather than fragmented actions. As shown in Figure 3, the QA pairs are balanced across 3 games---CS2 (651 QAs), Valorant (449), and OW2 (645)---to ensure robust evaluation across diverse visual rendering styles and game mechanics.

\paragraph{\textbf{Task Taxonomy.}}
As shown in Figure~\ref{fig:figure3_stat}, each QA pair is categorized under an orthogonal two-dimensional taxonomy of cognitive ability and expert-level esports knowledge. The taxonomy details are as follows:

\begin{itemize}[leftmargin=*, topsep=2pt]
\item \textbf{Cognitive Capability Axis:} This axis comprises 11 fine-grained task types organized into two hierarchical levels. The \textit{Perception Level} assesses a model’s ability to extract explicit visual information from high-velocity dynamics and information-dense interfaces, covering object recognition, egocentric action recognition, and interface, spatial, and temporal perception. The \textit{Reasoning Level} requires sophisticated logical deduction built upon robust visual perception, including spatial, temporal, individual action, and multi-agent interaction reasoning, as well as intent and counterfactual reasoning.
\item \textbf{Esports Knowledge Axis:} Questions are distributed across 6 specialized domains that demand esports knowledge and expert-level tactical understanding. These domains can be categorized into \textit{Micro-Level} operations (planned tactics, player mechanics, and adaptive coordination) and \textit{Macro-Level} progression (map control, match progression, and resource management).
\end{itemize}
This dual-axis design enables our benchmark to analyze model capabilities from both cognitive and application-oriented perspectives, facilitating a more comprehensive and meaningful evaluation.

\paragraph{\textbf{Quality Control.}}
To ensure the benchmark evaluates genuine multimodal reasoning rather than language pattern matching, we carefully controlled linguistic properties and choice distribution. The average length of a question is 14.9 words, with each option averaging 9.9 words and the ground-truth answer averaging 9.7 words. We incorporate professional esports terminology, such as "defaulting," "eco-rounds," and "flanking," to maintain technical granularity and require models to understand domain-specific concepts. Correct answers are strictly balanced across the four choices (A: 25.1\%, B: 25.0\%, C: 25.0\%, D: 24.9\%). Additionally, through anti-leakage processing, we neutralized potential textual shortcuts, ensuring the task cannot be solved via linguistic priors or dataset common sense.


\section{Experiments}

\begin{table*}[t]
\centering
\resizebox{\textwidth}{!}{
\begin{tabular}{l | c | *{3}{wc{1.2cm} wc{1.2cm}} ccc} 
\toprule
\multirow{2}{*}{\textbf{Models}} & \multirow{2}{*}{\textbf{Params}} & \multicolumn{2}{c}{\textbf{Counter-Strike 2 (\%)}} & \multicolumn{2}{c}{\textbf{Valorant (\%)}} & \multicolumn{2}{c}{\textbf{Overwatch 2 (\%)}} & \multicolumn{3}{c}{\textbf{Overall (\%)}} \\
\cmidrule(lr){3-4} \cmidrule(lr){5-6} \cmidrule(lr){7-8} \cmidrule(lr){9-11}
 &  & \textit{Perc.} & \textit{Reas.} & \textit{Perc.} & \textit{Reas.} & \textit{Perc.} & \textit{Reas.} & \textit{Perc.} & \textit{Reas.} & \textit{Total} \\
\midrule
\multicolumn{11}{c}{\textit{Closed-source Video-LLMs}} \\
\midrule
Gemini 3 Flash~\citeyearpar{pichai2025gemini3}  & - & 70.09 & 43.33 & 47.33 & 33.69 & 50.54 & 42.91 & 56.66 & 40.81 & 49.74 \\
GPT-5~\citeyearpar{openai2025gpt5} & - & 82.05 & 66.00 & 71.76 & 70.05 & 68.92 & 68.73 & 74.36 & 67.98 & \textbf{71.58} \\
Claude-Sonnet-4.5~\citeyearpar{anthropic2025sonnet45} & - & 65.24 & 45.33 & 54.20 & 50.80 & 46.76 & 46.91 & 55.34 & 47.24 & 51.81 \\
Doubao-Seed-1.8~\citeyearpar{bytedance2025seed18} & - & 73.79 & 56.33 & 61.45 & 55.61 & 60.81 & 55.64 & 65.62 & 55.91 & 61.38 \\
\midrule
\multicolumn{11}{c}{\textit{Open-source Video-LLMs}} \\
\midrule
Qwen3-VL~\citeyearpar{bai2025qwen3}   & 8B & 68.66 & 42.00 & 45.42 & 47.06 & 50.27 & 42.55 & 55.54 & 43.44 & \textbf{50.26} \\
InternVL-3.5~\citeyearpar{wang2025internvl35} & 8B & 45.58 & 35.33 & 37.40 & 41.18 & 36.22 & 30.91 & 39.88 & 35.17 & 37.82 \\
LLaVA-NeXT-Video~\citeyearpar{li2024llavanextvideo} & 7B & 25.07 & 31.33 & 27.48 & 28.88 & 24.32 & 21.45 & 25.43 & 27.17 & 26.19 \\
LLaVA-OneVision~\citeyearpar{li2025llavaonevision} & 7B & 37.61 & 32.33 & 30.53 & 35.83 & 33.51 & 30.91 & 34.18 & 32.68 & 33.52 \\
EgoGPT~\citeyearpar{yang2025egolife} & 7B & 34.19 & 34.67 & 30.53 & 39.04 & 36.76 & 33.82 & 34.18 & 35.43 & 34.73 \\
\bottomrule
\end{tabular}
}
\caption{Main evaluation results of 9 Video-LLMs on EgoEsportsQA. \textit{Perc.}: Perception Level; \textit{Reas.}: Reasoning Level.}
\label{tab:main_result}
\vspace{-0.5cm}
\end{table*}

In this section, we evaluate a wide range of Video-LLMs on our EgoEsportsQA benchmark, covering experimental settings, quantitative results for multiple Video-LLMs across two dimensions, as well as rich ablation studies.

\subsection{Settings}


\paragraph{\textbf{Models.}}
We systematically evaluate 9 representative Video-LLMs on EgoEsportsQA. For proprietary models, we include Gemini 3 Flash~\cite{pichai2025gemini3}, GPT-5~\cite{openai2025gpt5}, Claude-Sonnet-4.5~\cite{anthropic2025sonnet45}, and Doubao-Seed-1.8~\cite{bytedance2025seed18}. For open-source Video-LLMs, we evaluate Qwen3-VL-8B-Instruct~\cite{bai2025qwen3}, InternVL-3.5-8B-Instruct~\cite{wang2025internvl35}, LLaVA-NeXT-Video~\cite{li2024llavanextvideo}, and LLaVA-OneVision~\cite{li2025llavaonevision}. Additionally, we evaluate EgoGPT~\cite{yang2025egolife} (a variant of LLaVA-OneVision fine-tuned on real-world egocentric video datasets) to explore the impact of cross-domain adaptation. Due to its native audio support and cost-efficiency, Gemini 3 Flash is also utilized as the primary model for our ablation studies.

\paragraph{\textbf{Video Input.}}
In experiments where not otherwise specified, we sample frames at 1 fps to balance temporal coverage and computational cost. Since the maximum video duration in our dataset is 196 seconds, some models cannot accommodate the full video at 1 fps due to their input frame limits; in such cases, we adopt a uniform sampling strategy to meet their maximum capacity. Specifically, GPT-5 supports 50 frames, Claude-Sonnet-4.5 supports 100 frames, Doubao-Seed-1.8 supports 92 frames, InternVL-3.5 supports 64 frames, and both LLaVA-NeXT-Video and LLaVA-OneVision support 32 frames. All frames are resized to a fixed 720p (1280 $\times$ 720) resolution by default to preserve the clarity of UI elements essential for information extraction.


\paragraph{\textbf{Evaluation Protocol.}}
Since all questions follow a multiple-choice format with four options, we adopt \textit{Accuracy} as the primary evaluation metric, computed by directly comparing the model’s output with the ground-truth answer without relying on external models.

\subsection{Main Results}

\begin{figure}[h] 
    \centering
    
    \begin{subfigure}{\linewidth}
        \centering
        \includegraphics[width=0.9\linewidth]{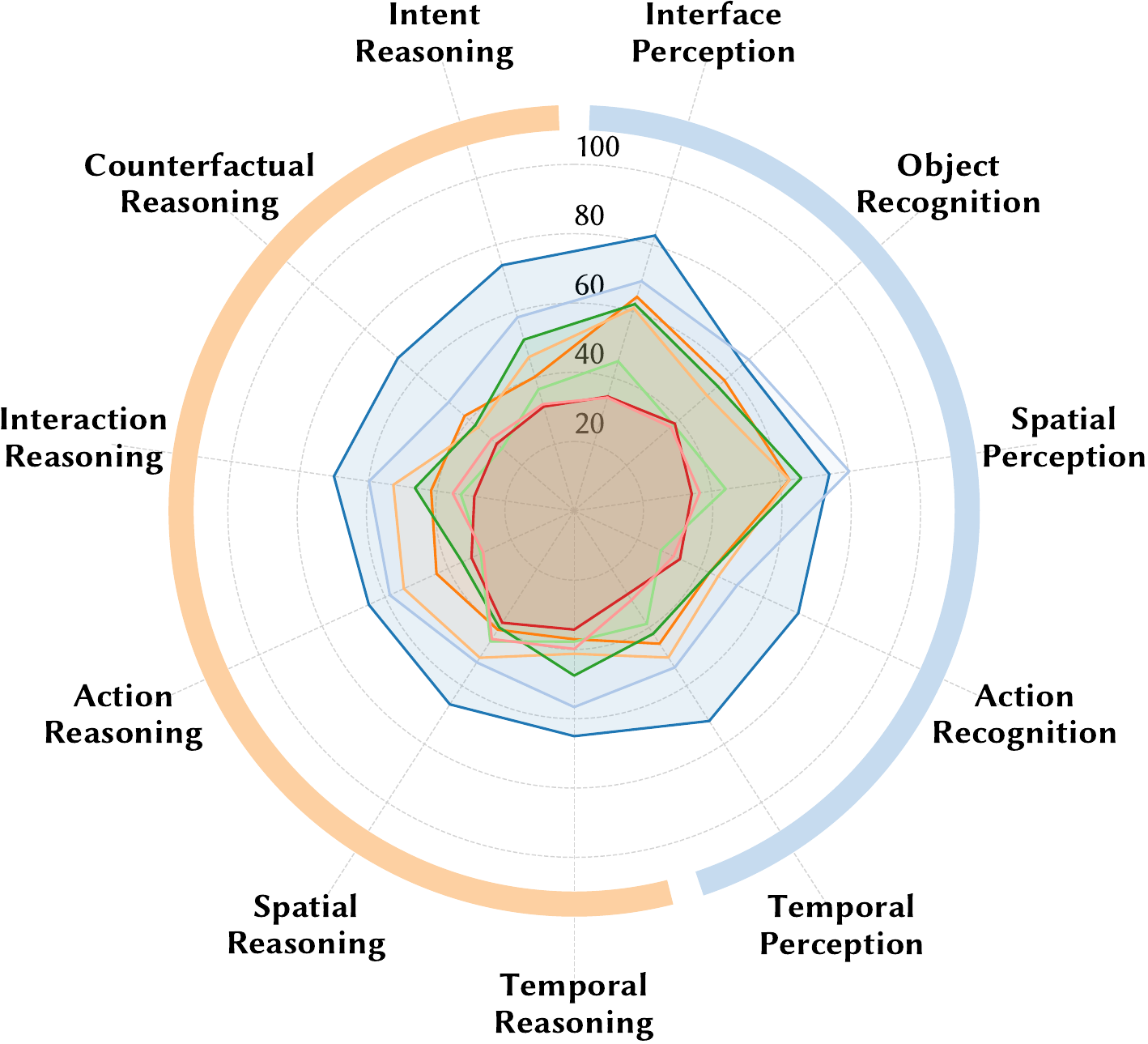}
        \label{fig:cognitive_4a}
    \end{subfigure}
    \vspace{0cm} 
    
    \begin{subfigure}{\linewidth}
        \centering
        \includegraphics[width=0.9\linewidth]{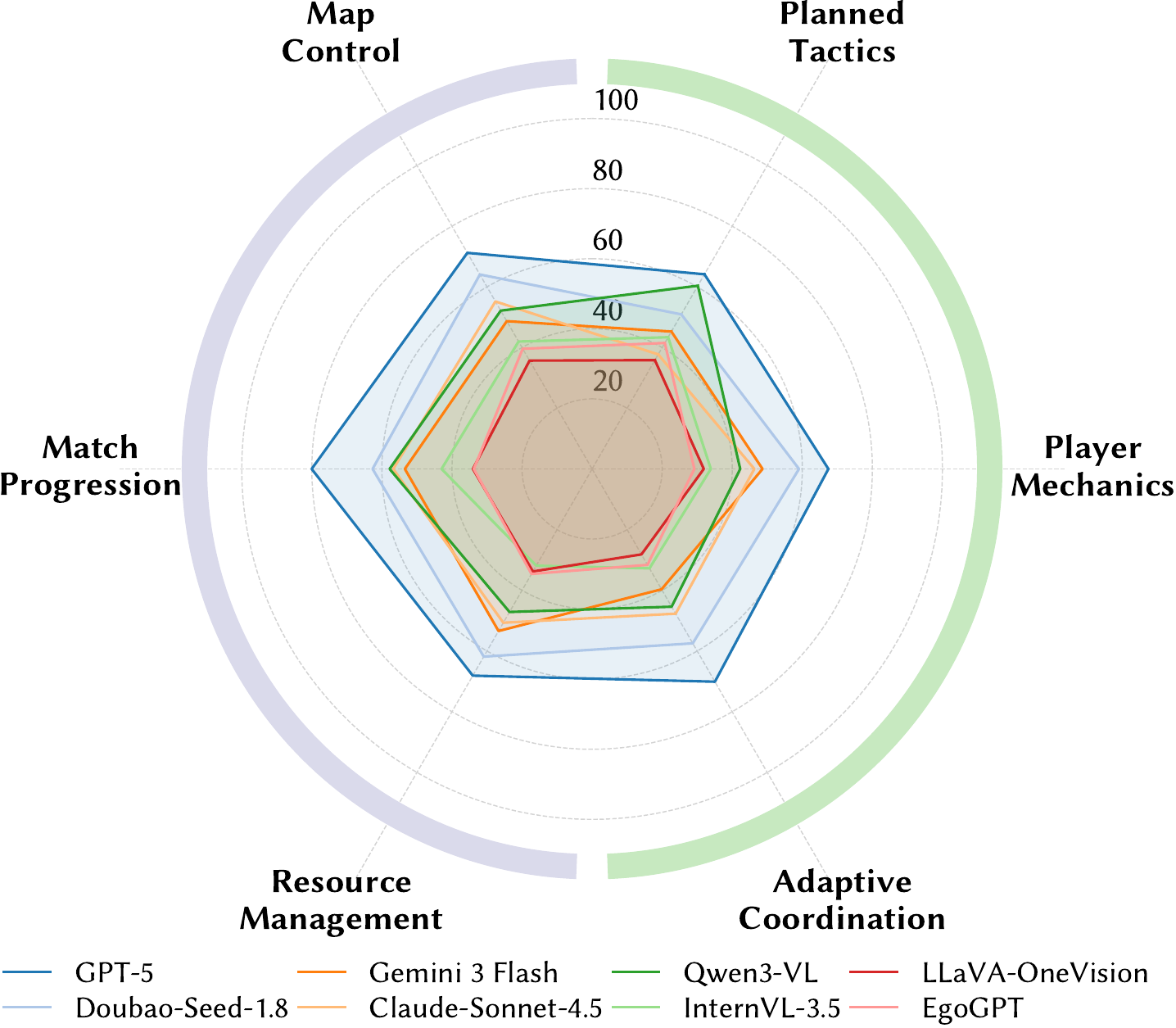}
        \label{fig:esports_4b}
    \end{subfigure}
    
    \caption{Performance breakdown of 8 Video-LLMs across the cognitive capability and esports knowledge dimensions.}
    \label{fig:figure4_2axes}
    \vspace{-0.5cm}
\end{figure}

The overall evaluation results on the EgoEsportsQA benchmark are summarized in Table~\ref{tab:main_result} (which reports the accuracy across 3 games and decouples performance into perception and reasoning levels) and Figure~\ref{fig:figure4_2axes} (which illustrates the performance of multiple models across two orthogonal dimensions). Two critical observations are as follows:

\paragraph{\textbf{Cognitive Capability: Perception vs. Reasoning.}}
The first major dimension assesses the fundamental visual intelligence of Video-LLMs. Overall, interpreting high-velocity esports videos remains a formidable challenge, with considerable room for improvement across all cognitive sub-tasks. Even the state-of-the-art closed-source model, GPT-5~\cite{openai2025gpt5}, achieves only 74.36\% in perception and 67.98\% in reasoning. A consistent trend across all evaluated models is that \textit{perception serves as the foundation, while reasoning forms the primary bottleneck} (see Figure~\ref{fig:figure4_2axes}). Models generally struggle to deduce tactical intents or counterfactuals even when explicit visual cues are successfully extracted. Additionally, as shown in Table~\ref{tab:main_result}, a substantial performance gap ($\sim$21\%) exists between proprietary and open-source models, indicating that open-source architectures still lack robust multimodal alignment for complex virtual egocentric environments.

Our evaluation also reveals clear performance disparities across different FPS titles, directly correlating with their \textit{distinct visual characteristics}. Models consistently perform best on CS2, which features a relatively photorealistic art style and intuitive visual physics closely resembling the real world. Valorant, adopting a stylized, semi-cartoonish aesthetic with distinct ability visual effects, poses a medium difficulty. In addition, OW2 yields the lowest accuracy across the board; its high-speed, chaotic screen motion combined with an extremely cluttered, information-dense UI poses a great challenge for current Video-LLMs.


Furthermore, although evaluating EgoGPT shows that the visual domain gap between real and virtual environments brings no accuracy gain in perception, its internalized first-person logical priors still benefit reasoning~\cite{yang2025egolife}. This suggests that egocentric reasoning logic is somewhat transferable, while visual perception remains domain-dependent.

\paragraph{\textbf{Esports Knowledge: Macro-Progression vs. Micro-Operation.}}
The second major dimension evaluates the models' grasp of domain-specific expertise, which is crucial for deploying Video-LLMs in real-world esports applications.

Our analysis reveals that current models exhibit significantly \textit{better proficiency in macro-level progression compared to micro-level operations}. As shown in Figure~\ref{fig:figure4_2axes}, tasks categorized under map control, match progression, and resource management yield higher average accuracies. This indicates that Video-LLMs are relatively adept at understanding the overall game state, likely because macro-strategies depend on global visual layouts and logic commonly found in their text-training corpora.

In contrast, performance decreases considerably on micro-level categories, including planned tactics, adaptive coordination, and player mechanics. These sub-domains require the model to capture split-second, fine-grained mechanical executions. The difficulty of extracting such transient, pixel-level interactions from compressed video tokens highlights a critical limitation: while Video-LLMs can reasonably infer of what the overall game situation is, they face profound difficulties in perceiving and interpreting how the players precisely execute actions. This limitation also represents a key bottleneck for applications such as esports commentary and visual agents.

\subsection{Ablation Studies}
\label{subsec:ablation}

To thoroughly investigate the benchmark's properties and diagnose the architectural bottlenecks of current Video-LLMs, we conduct extensive ablation studies on several key factors: input modalities, frame sampling rate, input video resolution, and the scope of the temporal context window.

\paragraph{\textbf{Impact of Input Modalities.}}
To understand the reliance of EgoEsportsQA on specific sensory inputs and evaluate cross-modal synergy, we ablate the input modalities, as shown in Table~\ref{tab:ablation_modality}. In the text-only setting on Gemini 3 Flash, overall performance drops to 24.47\%, virtually equivalent to random guessing. This confirms that models cannot bypass visual observation by relying solely on language priors. Incorporating the visual modality brings substantial improvement, pushing the score to 49.47\%. Notably, the further addition of the audio modality yields a profound enhancement: while perception accuracy remains relatively stable, reasoning accuracy surges from 40.81\% to 49.74\%. This pattern closely mirrors the cognitive mechanics of FPS esports, where \textit{auditory cues reveal critical hidden-state information} that is indispensable for high-level tactical deduction and intent prediction (e.g., directional footsteps and ability voice lines).


\begin{table}[t]
\centering
\resizebox{\columnwidth}{!}{
\begin{tabular}{ccc | ccc} 
\toprule
\multicolumn{3}{c |}{\textbf{Modality}} & \multicolumn{3}{c}{\textbf{Overall Score (\%)}} \\
\cmidrule(lr){1-3} \cmidrule(lr){4-6}
Text & Visual & Audio & \textit{Perception} & \textit{Reasoning} & \textit{Total} \\
\midrule
\ding{51} &  \ding{55}          &     \ding{55}       & 24.01 & 25.07 & 24.47 \\
\ding{51} & \ding{51} &  \ding{55}          & 56.66 & 40.81 & 49.47 \\
\ding{51} & \ding{51} & \ding{51} & \textbf{57.27}  & \textbf{49.74}     & \textbf{53.98}     \\
\bottomrule
\end{tabular}
}
\caption{Ablation study on different input modalities (Text, Visual, Audio). ``Visual'' refers to image sequences.}
\label{tab:ablation_modality}
\end{table}


\begin{table}[t]
\centering
\resizebox{0.9\columnwidth}{!}{
\begin{tabular}{c | ccc}
\toprule
\textbf{Sampling Rate} & \textbf{Perception} & \textbf{Reasoning} & \textbf{Total} \\
\midrule
0.5 fps          & 49.24               & 38.58              & 44.58            \\
1.0 fps            & \textbf{56.66}               & \textbf{40.81}              & \textbf{49.74}            \\
2.0 fps            & 46.19               & 39.11               & 43.09        \\
\bottomrule
\end{tabular}}
\caption{Ablation study on video frame sampling rates.}
\label{tab:ablation_fps}
\vspace{-0.5cm}
\end{table}

\paragraph{\textbf{Impact of Frame Sampling Rate.}}
We ablate the frame extraction rate to identify the optimal visual density for high-speed esports environments, as shown in Table~\ref{tab:ablation_fps}. Extracting frames at 1 fps achieves the best balance, yielding 49.74\% overall accuracy. Reducing the sampling rate to 0.5 fps degrades total performance to 44.58\%: given the high-velocity nature of FPS games, \textit{sparse temporal sampling inevitably misses fleeting yet critical visual cues}. More intriguingly, doubling the temporal resolution to 2 fps also degrades perception accuracy to 46.19\%. We attribute this performance drop to \textit{contextual overload and attention dilution}---processing overly dense frame sequences exponentially increases the number of visual tokens, which overwhelms the Video-LLM's context window and causes its attention mechanism to lose focus on salient details.


\begin{table}[t]
\centering
\resizebox{0.9\columnwidth}{!}{
\begin{tabular}{c|ccc}
\toprule
\textbf{Resolution} & \textbf{Perception} & \textbf{Reasoning} & \textbf{Total} \\
\midrule
 256 $\times$ 144      &     30.52                &     30.05               &     30.32           \\
640 $\times$ 360      &        47.41             &     36.48               &      42.64          \\
 1280 $\times$ 720     &      \textbf{56.66}               &     \textbf{40.81}               &    \textbf{49.74}            \\
 1920 $\times$ 1080   &       50.76              &     36.48               &     44.53         \\
\bottomrule
\end{tabular}
}
\caption{Ablation study on input spatial resolutions.}
\label{tab:ablation_resolution}
\end{table}

\paragraph{\textbf{Impact of Spatial Resolution.}}
Table~\ref{tab:ablation_resolution} highlights the severe impact of visual clarity. The 720p resolution emerges as the sweet spot, yielding the highest scores. Extreme spatial down-sampling to 144p or 360p causes overall performance to collapse (to 30.32\% and 42.64\%, respectively), primarily due to severe \textit{UI blindness} where crucial numbers and minimap icons become entirely illegible. Interestingly, increasing the resolution to 1080p decreases the total score to 44.53\%. The performance degradation is likely caused by \textit{excessive token compression, spatial slicing, and positional embedding mismatch under fixed context budgets}~\cite{zhang2025q, huang2025hires}, which may erase fine-grained details and break spatial continuity.


\begin{table}[t]
\centering
\resizebox{\columnwidth}{!}{

\begin{tabular}{c|ccc}
\toprule
\textbf{Context Window} & \textbf{Perception}  & \textbf{Reasoning}   & \textbf{Total}       \\
\midrule
Local ($[t_{begin}, t_{end}]$)           &      62.36                &      46.85                &       55.59               \\
Expanded ($\pm 5$s margin)         & \textbf{63.48} & \textbf{46.98} & \textbf{56.28} \\
Global (Full Video)          & 56.66                & 40.81                & 49.74 \\
\bottomrule
\end{tabular}}
\caption{Ablation study on the scope of temporal context windows.}
\label{tab:ablation_window}
\vspace{-0.5cm}
\end{table}

\paragraph{\textbf{Impact of Temporal Context Window.}}
Finally, we ablate the temporal context provided around the queried event. For a question $\tilde{Q}$ with a time anchor denoted as $[t_{begin}, t_{end}]$ (where $t_{begin}$ may equal $t_{end}$, with an average duration of 10.2 seconds across all questions), we evaluate three context settings, all sampled at 1 fps: 1) \textit{Local Context}, which restricts frames strictly within $[t_{begin}, t_{end}]$; 2) \textit{Expanded Context}, which introduces a temporal padding by sampling within $[\max(0, t_{begin}-5), \min(t_N, t_{end}+5)]$; and 3) \textit{Global Context}, which processes the entire video. As shown in Table~\ref{tab:ablation_window}, the expanded context achieves the highest total accuracy (56.28\%), slightly outperforming the local context. This improvement indicates that expert-level perception and understanding benefits from \textit{observing the causal chain around the core event}, rather than an isolated, truncated window. Notably, the global context achieves the lowest overall performance score. Irrelevant frames introduce visual noise, which disperses the model’s attention and impairs accurate temporal grounding. This demonstrates that \textit{uncurated long-form videos readily overwhelm existing multimodal encoders}~\cite{liu2025bolt, yu2025framevoyager}. Overall, our long-video benchmark poses a considerable challenge for models to perform effective and robust attention allocation.



\section{Further Analysis and Discussion}
\label{sec:discussion}

Beyond standard benchmarking, we conduct further analyses to uncover the broader implications of EgoEsportsQA for multimodal research. Specifically, we explore cross-domain generalization between real and virtual egocentric environments, and validate our benchmark as a diagnostic proxy for downstream applications.

\subsection{Cross-Domain Egocentric Transfer}

A key question in egocentric vision is whether cognitive patterns learned in real-world environments transfer to virtual domains, and vice versa. Our main results (Table~\ref{tab:main_result}) already suggest a \textbf{real-to-virtual} transfer: EgoGPT, fine-tuned on large-scale and highly diverse real-life egocentric data~\cite{yang2025egolife}, shows stronger logical reasoning performance than the base model. This demonstrates that real-world egocentric data provides transferable structural priors by teaching the model general cognitive and visual reasoning patterns of the egocentric domain that remain valid under the visual shifts of virtual environments.

To explore the \textbf{virtual-to-real} transfer, we conduct a dedicated fine-tuning experiment. Specifically, we fine-tune the Qwen3-VL-8B~\cite{bai2025qwen3} model using 1,407 additionally constructed QA pairs generated via our pipeline on CS2, chosen for its high photorealism. We employ QLoRA~\cite{dettmers2023qlora} with a rank of 128 and a learning rate of $2e^{-5}$, training for 2 epochs on 4 NVIDIA A6000 GPUs. As shown in Table~\ref{tab:transferability}, the fine-tuned model achieves a performance boost on EgoEsportsQA CS2 subset, improving the accuracy from 56.37\% to 58.68\%, with a notable enhancement in reasoning.

More intriguingly, we apply our fine-tuned model to MVBench’s \textbf{Egocentric Navigation} task~\cite{li2024mvbench}, a challenging embodied cognitive task. It requires models to understand first-person video and language instructions to predict the next action without a global map, which traditional Video-LLMs struggle with due to the need for fine-grained spatio-temporal alignment and ego-motion understanding. Despite these difficulties, our model, fine-tuned \textbf{only on virtual egocentric data}, achieves a clear out-of-domain improvement, raising accuracy from 33.50\% to 36.00\%. Our training data account for only $\sim$0.176\% of Qwen3-VL’s multimodal training data ($\sim$800K), but already yield notable gains. As virtual esports data is \textbf{easier to collect}, scaling up such efficient data is expected to bring further improvements, enabling broader applications in real-world egocentric scenarios such as embodied intelligence and egocentric navigation.

\begin{table}[t]
\centering
\resizebox{\columnwidth}{!}{
\begin{tabular}{l | ccc c} 
\toprule
\multirow{2}{*}{\textbf{Model}} & \multicolumn{3}{c}{\textbf{EgoEsportsQA-CS2 (\%)}} & \textbf{MVBench (\%)} \\
\cmidrule(lr){2-4} \cmidrule(lr){5-5}
 & \textit{Perc.} & \textit{Reas.} & \textit{Total} & \textit{Ego. Nav.} \\ 
\midrule
Qwen3-VL-8B & 68.66 & 42.00 & 56.37 & 33.50 \\
\multicolumn{1}{r |}{+ CS2 Data} & \textbf{69.80} & \textbf{45.67} & \textbf{58.68} & \textbf{36.00} \\ 
\bottomrule
\end{tabular}
}
\caption{Performance of Qwen3-VL-8B before and after fine-tuning on CS2 data, demonstrating virtual-to-real transferability on MVBench Egocentric Navigation (\textit{Ego. Nav.}) task.}
\label{tab:transferability}
\vspace{-0.5cm}
\end{table}

\subsection{Proxy for Downstream Applications}

Beyond QA accuracy, we also explore whether performance on EgoEsportsQA can act as a preliminary indicator for models’ capabilities in downstream tasks, thus forging a connection to real applications. We hypothesize that models unable to achieve high scores on our benchmark would likely struggle in practical applications.

To investigate this potential correlation, we conduct a \textbf{Commentary Generation} case study. We carefully select 50 egocentric clips of 12 seconds in duration from professional CS2 matches and prompt 4 models spanning distinct performance tiers on the leaderboard (Table~\ref{tab:main_result})---GPT-5~\cite{openai2025gpt5}, Gemini 3 Flash~\cite{pichai2025gemini3}, InternVL-3.5~\cite{wang2025internvl35}, and LLaVA-OneVision~\cite{li2025llavaonevision}---to generate live commentary. Following an LLM-as-a-judge paradigm~\cite{chen2025livecc,xu2026streamingvlm}, we use GPT-4o~\cite{hurst2024gpt4o} to evaluate win rates based on \textbf{semantic alignment} (consistency in meaning, details, and key events) and \textbf{stylistic consistency} (similarity in tone, wording, and structural flow) relative to human commentary groundtruth.

\begin{table}[t]
\centering

\resizebox{\columnwidth}{!}{%
\begin{tabular}{l | c c c c}
\toprule
\multirow{2}{*}{\diagbox[width=2.8cm]{\textbf{Model A}}{\textbf{Model B}}} 
 & \textbf{GPT-5} & \textbf{Gemini 3} & \textbf{InternVL} & \textbf{LLaVA-OV} \\
\cmidrule(lr){2-5}
 & \multicolumn{4}{c}{\textit{Win Rate of A vs. B (\%)}} \\
\midrule
\textbf{GPT-5}          & -     & 78.0  & 98.0  & 100.0 \\
\textbf{Gemini 3 Flash} & 22.0  & -     & 98.0  & 100.0 \\
\textbf{InternVL-3.5}    & 2.0   & 2.0   & -     & 62.0  \\
\textbf{LLaVA-OneVision}       & 0.0   & 0.0   & 38.0  & -     \\
\bottomrule
\end{tabular}%
}
\caption{Head-to-head win rates (\%) for commentary generation on 50 CS2 match clips, evaluated via LLM-as-a-judge (GPT-4o). Each cell represents the win rate of the row model against the column model.}
\label{tab:commentary_winrate}
\vspace{-0.5cm}
\end{table}


Table~\ref{tab:commentary_winrate} presents the head-to-head win rates, which exhibit a \textbf{consistent trend} mirroring our EgoEsportsQA results. GPT‑5, the top performer on our benchmark, is followed by Gemini 3 Flash, and GPT‑5 outperforms Gemini 3 Flash in 78\% of cases. These two proprietary models also show substantial advantages over open-source alternatives, achieving nearly 100\% win rates against the two open-source models. Although the open-source small models can also produce fluent text, they suffer from frequent hallucinations regarding visual details and game knowledge, which suggests that small open-source models still remain unsuitable for real-world deployment.



While this analysis focuses on a representative subset of models and clips, the preliminary results may suggest that EgoEsportsQA could \textbf{serve as a guiding benchmark} for deploying Video-LLMs in real-world downstream applications. By quantifying the alignment between QA performance and generative utility, it not only reveals key perception and reasoning shortcomings but also provides a roadmap for optimizing Video-LLMs before their integration into sophisticated, live esports ecosystems.
\section{Conclusion}

In this work, we introduce \textbf{EgoEsportsQA}, a pioneering benchmark for evaluating Video-LLMs in high-velocity, information-dense virtual environments. By curating 1,745 expert-level QA pairs via a scalable six-stage pipeline, we systematically decouple model performance into cognitive capabilities and specialized esports knowledge. Our experiments expose critical bottlenecks: while current Video-LLMs demonstrate basic visual perception, they struggle with deep tactical reasoning and fail to capture transient, pixel-level micro-operations compared to overall macro-strategies. Furthermore, we validate the benchmark's value through virtual-to-real egocentric transferability and its role as a diagnostic proxy for downstream applications. Ultimately, EgoEsportsQA highlights the limitations of current Video-LLMs in egocentric video understanding, serving as a vital foundation for developing next-generation agents capable of reasoning and acting in complex, dynamic worlds.





\bibliography{assets/reference}




\end{document}